%% file: Run_This_Example.tex
\documentclass[Alon2,ChapterTOCs,singlecolor,11pt]{Alon}

\usepackage{url}
\usepackage{graphicx}
\usepackage[english]{babel}
\usepackage{epigrafica}
\usepackage[LGR,OT1]{fontenc}
\usepackage{lmodern}
\usepackage{multicol}
\usepackage{changepage}
\usepackage{epstopdf}
\usepackage{orcidlink}
\usepackage{bbding} 
\usepackage{amssymb}
\usepackage{amsmath,amsthm}

\usepackage{booktabs}
\usepackage{multirow}
\usepackage{array}

\usepackage{makeidx}
\usepackage{enumitem}
\usepackage{version}
\usepackage{lipsum} 

\usepackage{tikz,xcolor}

\definecolor{lime}{HTML}{A6CE39}
\DeclareRobustCommand{\orcidicon}{%
	\begin{tikzpicture}
	\draw[lime, fill=lime] (0,0) 
	circle [radius=0.16] 
	node[white] {{\fontfamily{qag}\selectfont \tiny ID}};
	\draw[white, fill=white] (-0.0625,0.095) 
	circle [radius=0.007];
	\end{tikzpicture}
	\hspace{-2mm}
}

\foreach \x in {A, ..., Z}{%
	\expandafter\xdef\csname orcid\x\endcsname{\noexpand\href{https://orcid.org/\csname orcidauthor\x\endcsname}{\noexpand\orcidicon}}
}

\renewcommand\bibfont{\fontsize{10}{12}\selectfont}

\PassOptionsToPackage{hidelinks}{hyperref}
\usepackage{hyperref}
\makeatletter
\AtBeginDocument{\let\Hy@CheckDuplicatePageAnchor\relax}
\makeatother

\theoremstyle{plain}

\theoremstyle{definition}

\theoremstyle{remark}

\title{Taylor and Francis Book Chapter}

\begin{document}

\frontmatter

\maketitle 

\setcounter{page}{7} 
\makeatletter
\providecommand{\Hy@tocdestname}{}
\makeatother
\tableofcontents
\pagenumbering{arabic}


\include{chapter1/ch1}

\bibliographystyle{plain}
\bibliography{bibtex_example}


\end{document}

%% file: chapter1/ch1.tex
\chapterauthor{Muhammad Jawad Chowdhury\orcidlink{0009-0009-4129-546X}, Adiba Hasan\orcidlink{0009-0006-9072-4054}, Ishrak Hossain\orcidlink{0009-0004-1995-6122}, Shahriar Ivan\Envelope~\orcidlink{0009-0001-2801-3028}, and Sabbir Ahmed
\orcidlink{0000-0001-5928-4886}
}
{Department of Computer Science and Engineering, Islamic University of Technology, Gazipur, Bangladesh.\\
\{jawad, adibahasan, ishrakhossain, shahriarivan, sabbirahmed\}@iut-dhaka.edu
}
\let\cleardoublepage\clearpage


\pdfbookmark{Introduction}{chap:introduction}
\chapter{Beyond Accuracy: A Qualitative Analysis of Vision-Language Models for Hate Speech Detection in Memes}

\chaptermark{Beyond Accuracy: A Qualitative Analysis of Vision-Language Models for Hate Speech Detection in Memes}

\newpage
\chapterinitial Memes have turned out to be a powerful tool through which individuals share their ideas concerning contemporary social and political problems. Their anonymity, as well as their ability to go viral, make them a powerful medium for spreading hate. It remains very difficult to identify such complex and context-dependent hate speech. Although they display excellent performance on multimodal tasks, vision-language models (VLMs) tend to ignore context, irony, and other subtle cues that play a key role in identifying hateful memes. In this work, we present a qualitative analysis of four state-of-the-art VLMs: LLaVA-7B, Qwen-VL, GPT-4o mini, and Claude 3 Haiku. We evaluate these models under zero-shot and few-shot prompting to examine how contextual framing influences their outputs. Our analysis goes beyond simple classification accuracy and focuses on a qualitative evaluation of the models' generated justifications, providing a more in-depth understanding of their thought processes and constraints when dealing with hateful memes.


\section{Introduction}
As a form of cultural expression, memes are usually image-based content with brief and humorous text, extensively spread online. Though they are commonly used for entertainment, they can also be used to portray hate speech against individuals or a community, which has unfortunately become a very common issue in recent times \cite{gandhi2024hateSpeech}.
The growing societal concern around online hate has led to increasing efforts from both industry and academia to address this challenge  \cite{velioglu2020detecting,das2020detecting,ivan2024hate,kiela2020hateful}. This task becomes more challenging when the inherent context requires both image and textual reference \cite{Hermida2023}. In the case of multimodal data, hate speech may arise from the combined interpretation of text and image rather than from the individual modalities taken into consideration separately.
Each modal may seem innocuous separately, but when taken as a whole, it can convey an offensive or hateful message when considered together \cite{kumar2022hate,10418458}.

\begin{figure}[htbp]
    \centering
    \begin{minipage}[b]{0.3\textwidth}
        \includegraphics[width=\textwidth]{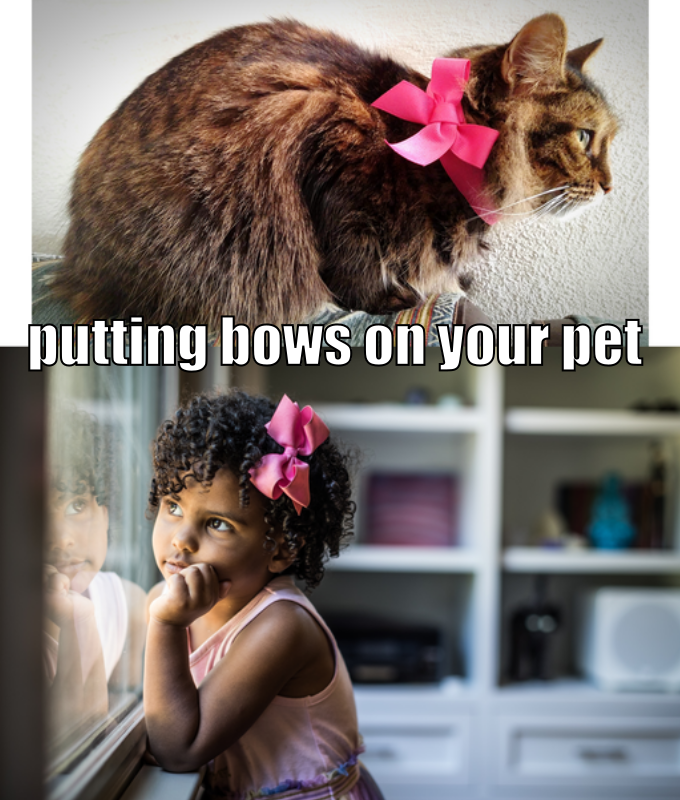}
    \end{minipage}
    \hfill
    \begin{minipage}[b]{0.3\textwidth}
        \includegraphics[width=\textwidth]{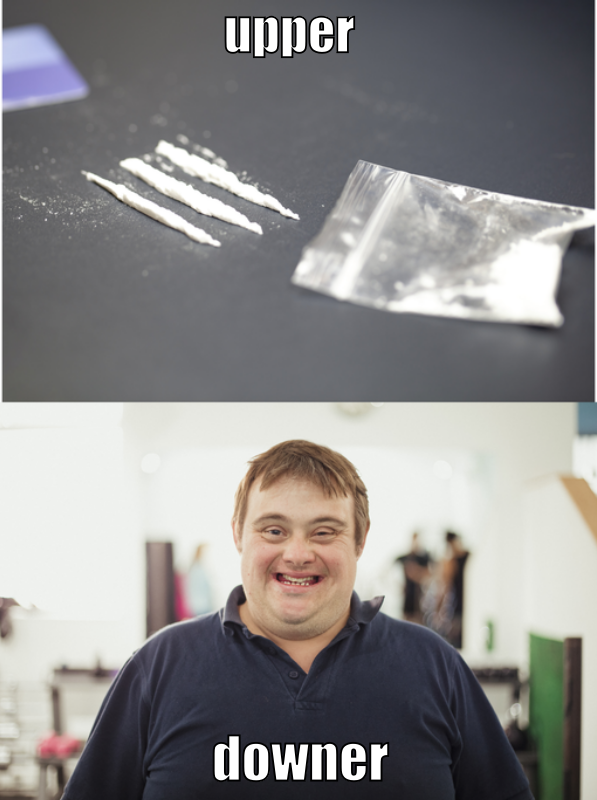}
    \end{minipage}
    \hfill
    \begin{minipage}[b]{0.3\textwidth}
        \includegraphics[width=\textwidth]{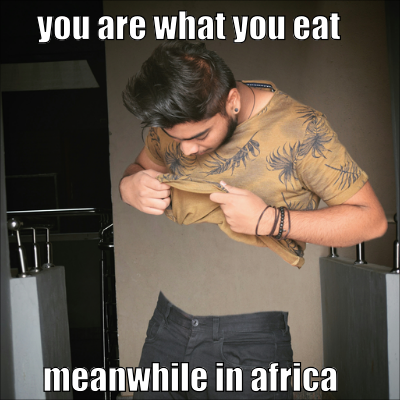}
    \end{minipage}
    \caption{Demonstration of sample meme images from \textbf{Hateful Memes Challenge Dataset (HMCD)}  \cite{kiela2020hateful}, where, without understanding the context, it is hard to capture the underlying hate speech towards a targeted community. Here, the memes are to be considered multimodally while understanding the context.}
    \label{fig:three-memes}
\end{figure}


Recent advances in deep learning have revolutionized the modern era by providing state-of-the-art solutions across a wide range of downstream tasks in computer vision \cite{ahmed2024exeNet,raiyan2025hasper,fuad2025aqua20,ahmed2022less}, natural language processing \cite{khan-etal-2023-banglachq,ridwan2023poem}, and related domains \cite{hasan2023GaitGCN,aziz2023banglaSER,khan2022rethinking,rahman2022twoDecades}. Building on these advances, modern language models have shown significant potential in terms of large-scale knowledge representation, contextual reasoning, and semantic structure capture.
\cite{ahmed2024Depression,alvi2023nervous,ahmed2021amIbeing,Ahmed2022}. This has led to the emergence of large language models (LLMs) \cite{kojima2022large} and vision-language models (VLMs)  \cite{ghosh2024exploring}, which extend these capabilities to multimodal understanding. LLMs and VLMs have shown strong performance in generating human-like text and understanding complex multimedia content  \cite{chen2023sharegpt4vimprovinglargemultimodal, arif2025hallucinaton, ajwad2025banglaCHQ, dai2023instructblip}. Since memes are inherently multimodal, often relying on sarcasm, cultural nuances, and intricate visual–textual interactions, they provide a particularly challenging testbed for assessing the real-world reasoning abilities of such models.

Recent advancements in multimodal learning have spurred interest in applying VLMs to this complex task of meme analysis  \cite{alayrac2022flamingo,ivan2024vision}. The Hateful Memes Challenge dataset proposed by Kiela \textit{et al.}  \cite{kiela2020hateful} introduced a benchmark, where non-hateful modalities may form hateful messages individually when interpreted together. Mathias \textit{et al.} \cite{lambert2021woah} extended this dataset by adding two sub-tasks where the objective was to identify the hatred categories in memes. Early research in this direction included using CLIP \cite{radford2021learning}, which maps images and text onto a shared embedding space. However, as a critical limitation, these systems, like HateCLIPper  \cite{kumar2022hate} and MemeCLIP  \cite{bikram2024memeclip}, often struggled with detecting covert or culturally nuanced hate, as they relied more on representation matching than deep contextual reasoning.

Modern pre-trained VLMs, like Flamingo  \cite{alayrac2022flamingo}, LLaVA  \cite{liu2023visual}, and GPT-4 \cite{openai2023gpt4} have recently gained significant popularity. Frameworks like \textit{MemeGuard} \cite{jha2024memeguard} used a fine-tuned VLM for interpreting meme contexts, while Hee \textit{et al.}  \cite{hee2023decoding} proposed the \textit{HatReD} dataset containing annotations of underlying hateful contextual reasons. Several recent research endeavors have focused on effective prompting strategies to guide VLM reasoning. Gavit \textit{et al.} \cite{gavit2025vlms,liu2025memeblip2} provided a structured analysis of VLM capabilities on various meme classification tasks, comparing methods like Zero-Shot learning \cite{tajwar2023improving}, Few-Shot learning \cite{ahmed2025dexnet,mehedi2024fslBHDR}, and Chain-of-Thought (CoT) \cite{wei2022chain}. Similarly, Zhuang \textit{et al.} \cite{inproceedings} used chain-of-thought prompting with GPT-4 to achieve state-of-the-art results, and Van and Wu \cite{van2025detecting} introduced a definition-guided prompting strategy.

In addition to benchmark performance, a good amount of research has been done on the common failure modes of these models, providing a foundation for our qualitative analysis \cite{ivan2024vision}. The central challenge of contextual misinterpretation, where models fail to synthesize multimodal cues, was a key motivator for the Hateful Memes Challenge itself  \cite{kiela2020hateful}. This is often compounded by models exhibiting an over-sensitivity to keywords, a form of unintended bias where the presence of a sensitive term can trigger a misclassification regardless of the benign context \cite{dixon2018measuring}. Furthermore, studies have highlighted a persistent weakness in detecting coded or nuanced hate, particularly when it involves irony or sarcasm, which models often interpret literally \cite{vidgen2019challenges}. Perhaps most insidiously, researchers have identified the phenomenon of models achieving correct classifications through entirely flawed reasoning— a `right for the wrong reasons' problem that masks a true lack of understanding \cite{niven2019probing}. These established challenges underscore the limitations of relying solely on accuracy metrics and motivate a deeper, more diagnostic approach to VLM evaluation \cite{bender2021dangers}.

While the literature has identified these critical failure modes, most empirical evaluations of modern VLMs still prioritize a narrow set of quantitative metrics or test a limited range of models. A critical gap remains in understanding how different model architectures (e.g., open-source and API-based) compare in their qualitative reasoning and failure modes under varied prompting conditions. Most of the existing works focus on classification accuracy, where the important factors in the decision-making process of different models were underexamined. To address this gap, our work provides the following key contributions:

\begin{itemize}[noitemsep]
    \item Rigorous empirical evaluation of four modern VLMs: LLaVA-7B, Qwen-VL, GPT-4o mini, and Claude 3 Haiku, to compare their capabilities.
    \item Systematic analysis on the impact of prompting strategies by comparing a zero-shot approach against a context-rich few-shot approach for each model, revealing their sensitivity to prompt design. 
    \item Detailed qualitative analysis demonstrating how established failure modes, such as flawed reasoning and contextual misinterpretation, manifest in these models.
\end{itemize}



\section{Methodology}

\begin{figure}[htbp]
    \centering
    \includegraphics[width=\textwidth]{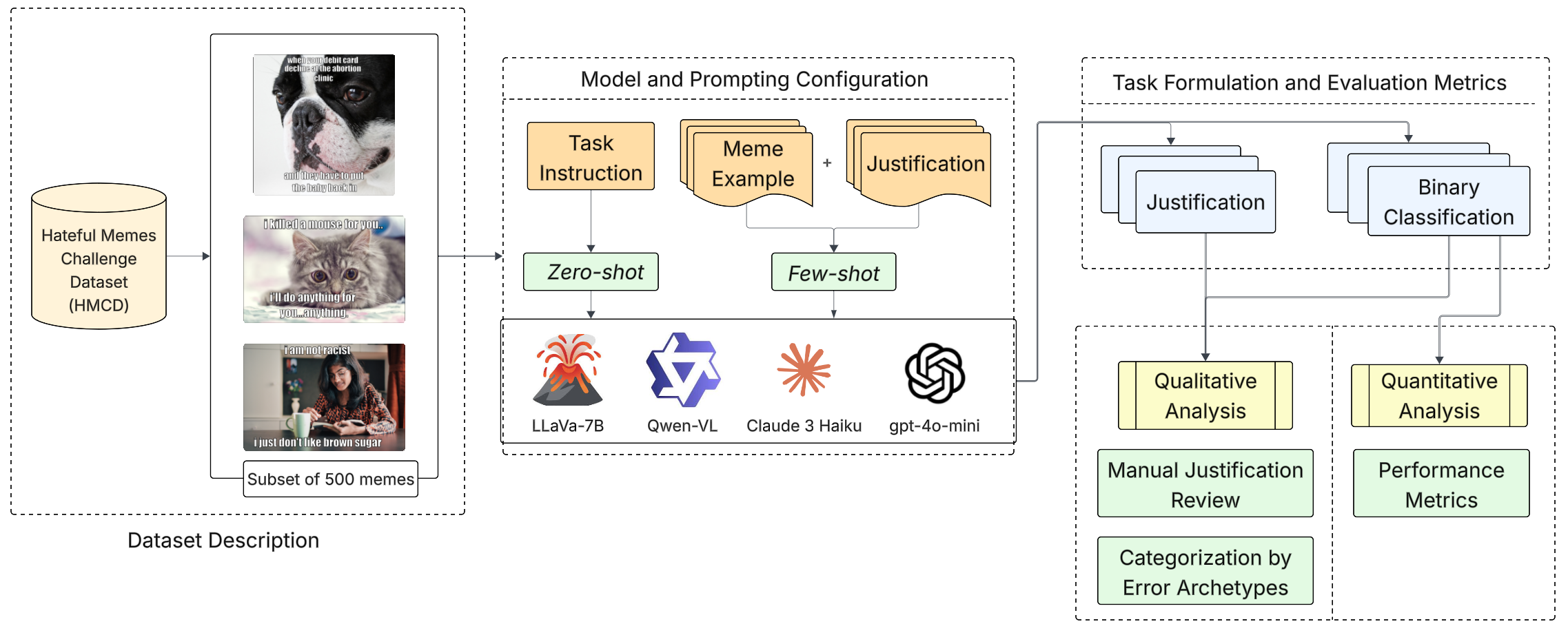}
    \caption{Overview of the empirical evaluation framework, which flows from dataset sampling and prompt configuration (zero-shot vs. few-shot) to evaluation across four VLMs, culminating in the analysis of their classification and justification outputs.}
    \label{fig:meme-example}
\end{figure}

To evaluate the effectiveness of modern VLMs in detecting hate speech within multimodal memes, we conducted an empirical study using four state-of-the-art models under two prompting strategies. Our focus was on qualitative analysis to capture the nuances of model reasoning and failure modes, rather than relying solely on quantitative metrics. Figure \ref{fig:meme-example} provides an overview of the approach, consisting of three components: 1) Dataset description, 2) Model and prompt settings, and 3) Task formulation with evaluation metrics supporting the analysis.

\subsection{Dataset}

We conducted our experiments using the \textbf{Hateful Memes Challenge Dataset (HMCD)} \cite{kiela2020hateful}, introduced in the NeurIPS 2020 Hateful Memes competition. As meme classification often depends on the joint interpretation of visual and textual content, this benchmark is specifically designed to evaluate complex multimodal reasoning. A key challenge of the dataset is the inclusion of \textit{benign confounders}, which are instances that appear hateful when viewed through a single modality but are non-hateful when both modalities are considered together.

The full dataset contains \textbf{10,000} image-text pairs labeled as either \textit{hateful} or \textit{non-hateful}. Each sample falls into one of five semantic categories \cite{kiela2020hateful}:
\begin{itemize}[noitemsep]
    \item \textbf{Random Non-Hateful}: Neutral memes without harmful content.
    \item \textbf{Benign Image Confounders}: Appear hateful based on the caption alone, but the image negates the hateful implication.
    \item \textbf{Benign Text Confounders}: Appear hateful from the image alone, but the caption clarifies the intent.
    \item \textbf{Unimodal Hate}: Meme is hateful based solely on either the image or the text.
    \item \textbf{Multimodal Hate}: Hatefulness emerges only when image and text are interpreted together.
\end{itemize}

We experimented with a subset of \textbf{500 memes} from the \texttt{dev\_seen.json} split. This subset is \textit{class-balanced}, consisting of \textbf{250 hateful} and \textbf{250 non-hateful} samples. The selected examples reflect the diversity and complexity of the original dataset, including instances from all five semantic categories. This is important as in this context of multimodal hate detection, understanding arises from analyzing both visual and textual content jointly. 
Using balanced data ensures fair comparisons across models and prompts.


\subsection{Model and Prompting Configurations}


We evaluated four large VLMs, chosen to provide a balanced perspective on their diverse reasoning capabilities. Our selection includes models known for high-precision visual grounding (Qwen2-7B)\cite{ahmed2025qwen}, strong general-purpose inference (GPT-4o mini \cite{openai2024gpt4o}, Claude 3 Haiku \cite{anthropic2024haiku}), and foundational instruction-following abilities (LLaVA-7B) \cite{liu2023visual}. This selection allows us to compare three distinct open-source models against a highly capable, proprietary API-based model (Claude 3 Haiku), which serves as a state-of-the-art performance benchmark. To assess the models’ sensitivity to prompt design, we use two distinct prompting strategies: \textit{zero-shot} and \textit{few-shot}, commonly used in evaluating reasoning capabilities of LLMs \cite{brown2020language}.

\begin{table}[t]
\centering
\renewcommand{\arraystretch}{1.5} 
\caption{Sample Zero-Shot Prompt for Meme Classification}
\vspace{0.5em}
\begin{tabular}{|p{0.9\linewidth}|}
\hline
\textbf{Prompt:} \\[0.1cm] \hline 
You are a content moderation AI. Your task is to classify a meme as either `hateful' or `non-hateful'.
Analyze both the image and the accompanying text. 
Respond in exactly this format (no explanation before or after): \\
\textbf{Classification:} hateful or non-hateful \\
\textbf{Justification:} \textless your brief reason\textgreater \\
\textbf{Example:} \\
\textbf{Classification:} non-hateful \\
\textbf{Justification:} The meme contains no offensive content and appears neutral. \\ \hline
\end{tabular}
\end{table}

In the \textbf{zero-shot} setting, models are provided only with task instructions to classify a meme as \texttt{hateful} or \texttt{non-hateful}, accompanied by a justification, with no examples. In contrast, the \textbf{few-shot} prompt presents three labeled meme examples with justifications before the target instance. This design enables in-context learning and evaluates whether exposure to examples enhances predictive accuracy and reasoning consistency.

\subsection{Task Formulation and Evaluation}
The task is formulated as a binary classification problem. Let our dataset be $D = \{(M_i, y_i)\}_{i=1}^N$, where each sample consists of a meme $M_i = (I_i, T_i)$ composed of an image $I_i$ and its associated text $T_i$. The ground truth label is $y_i \in \{0, 1\}$. A Vision Language Model, defined as a function $f_\theta$, analyzes the meme according to the given prompt $P_j$ (where $j \in \{\text{zero}, \text{few}\}$) to produce a prediction:
\[
f_\theta(I_i, T_i, P_j) \rightarrow (\hat{y}_{ij}, J_{ij})
\]
where $\hat{y}_{ij} \in \{0, 1\}$ is the predicted label and $J_{ij}$ is the generated text justification. Each model-prompt configuration is evaluated on the same set of 500 memes.


We performed a deep qualitative analysis of the models' reasoning capabilities, which involved a manual review of the misclassified samples for each model-prompt configuration, with a specific focus on the textual justification ($J_{ij}$) provided by the model. This analysis moves beyond simply identifying \textit{if} a model was wrong to understanding \textit{why} it was wrong.  

The primary goal of this analysis is to identify recurring error archetypes and diagnose the underlying failure modes in the models' reasoning. Our manual review focused on categorizing these failures, looking for patterns such as:
\begin{itemize}[noitemsep]

    \item \textbf{Prompt Sensitivity and Reasoning Failure:} Investigates how a model's reasoning pathway and final decision are influenced by the prompt design \cite{zhao2021calibrate}.
    
    \item \textbf{Correct Classification But Flawed Reasoning :} Instances where the model produces the correct classification label, but its justification reveals a flawed, irrelevant, or logically unsound reasoning process \cite{niven2019probing}.
    
    
    
    \item \textbf{Contextual Misinterpretation:} Instances where the model correctly identifies individual elements (e.g., objects in an image, keywords in text) but fails to grasp the contextual meaning that makes the meme hateful or benign \cite{kiela2020hateful}.
    
    \item \textbf{Over-sensitivity to Keywords:} Errors where a model incorrectly flags benign content as hateful due to the presence of sensitive keywords (e.g., related to race, religion, or gender), even when used in a non-hateful context \cite{dixon2018measuring}.
    
    \item \textbf{Failure to Detect Coded or Nuanced Hate:} Cases where models miss hate speech that relies on subtle stereotypes, cultural in-jokes, or coded language, rather than explicit slurs \cite{vidgen2019challenges}. This is particularly relevant for the `benign confounder' memes in the HMCD.
    
\end{itemize}




\vspace{-4mm}


\section{Empirical Analysis and Findings}


Understanding model behavior in \textit{multimodal classification tasks} requires more than just numerical metrics. To this end, we conduct a multi-faceted analysis of four VLMs on hateful meme understanding. Through manual inspection of model outputs, we can find patterns in both correct and incorrect classifications, as well as recurring inconsistencies in generated justifications.  Finally, we do a quantitative evaluation, where we report standard performance metrics to see how well the overall classification accuracy and robustness hold up across all model-prompt configurations.


\subsection{Qualitative Error Analysis}

Although quantitative measurements offer a broad perspective, a qualitative examination of misclassified memes is crucial for comprehending the particular failure modes of these sophisticated VLMs. By reviewing the justifications provided by the models for their incorrect predictions, we identified several recurring error archetypes.

\subsubsection{Prompt Sensitivity and Reasoning Failure}

\begin{figure}[h!]
    \centering
    \includegraphics[width=.97\textwidth]{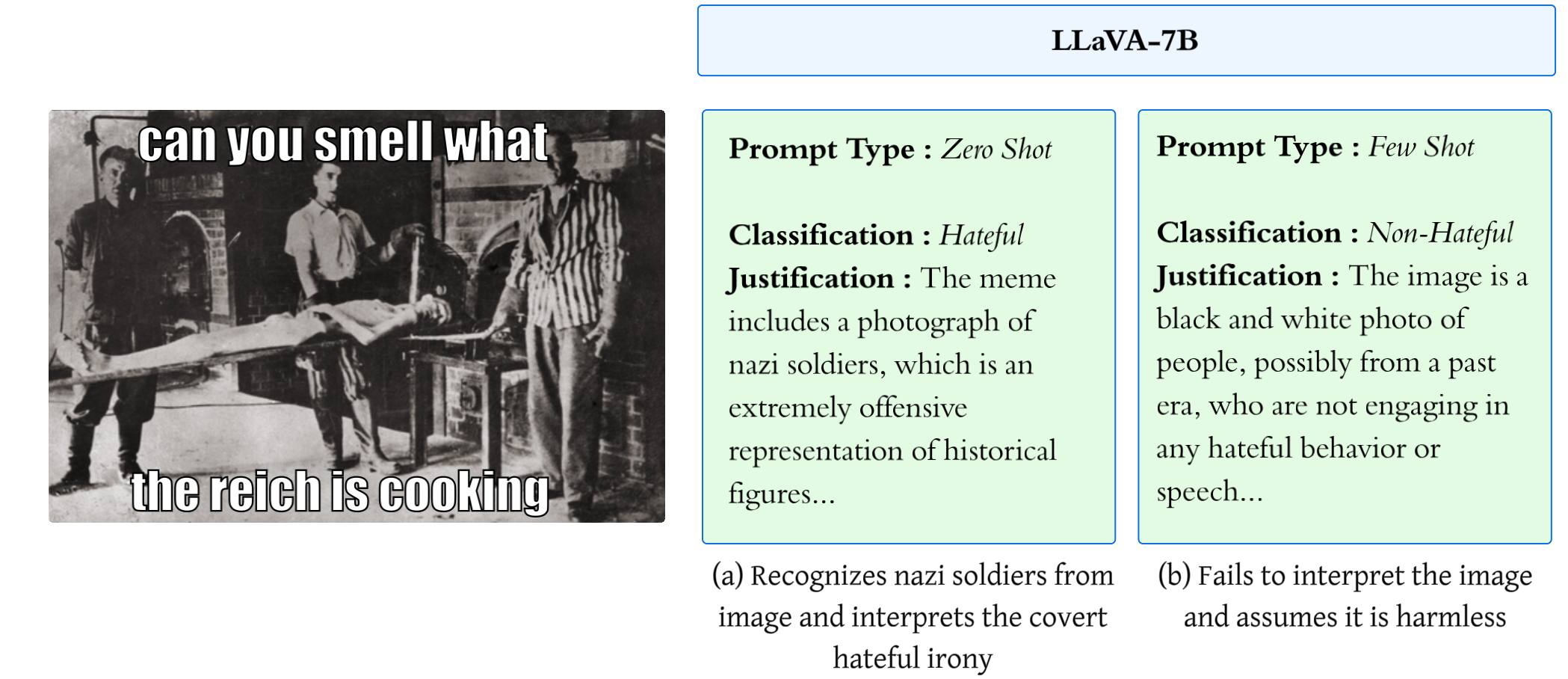} 
    \caption{For certain models, non-representative few-shot examples can be more detrimental than a zero-shot prompt, as they can suppress the model’s ability to apply its own  knowledge}
    \label{fig:Intra-model}
\end{figure}

A predominant failure mode identified was contextual misunderstanding, when a model's reasoning was significantly affected by the prompt. An illustrative instance is LLaVA-7B’s examination of the meme in Figure ~\ref{fig:Intra-model}, which presents a horrible humor alluding to the Holocaust and provoked radically conflicting reactions from the model and the model's classification transitioned from \texttt{non-hateful} to \texttt{hateful} upon altering the prompt from a zero-shot to a few-shot instruction.

\textit{Analysis}: This drastic shift reveals a critical vulnerability we term \textbf{Example-Induced Bias}, a phenomenon related to the known sensitivity of large models to in-context examples \cite{workrethinking}. In the zero-shot setting, the model's prediction $\hat{y}$ for a meme $M$ is primarily conditioned on its vast internal knowledge base, $\mathcal{K}$, following $P(\hat{y} | M, \mathcal{K})$. 
In the few-shot setting, however, the decision becomes heavily conditioned on the small set of provided examples, $\mathcal{E}$. The model's behavior suggests that it heavily discounts its internal knowledge. Ideally, the model should integrate both sources of information, but the observed failure can be expressed as the model's reasoning collapsing from the ideal to the biased state:
\[
P(\hat{y} | M, \mathcal{K}, \mathcal{E}) \rightarrow P(\hat{y} | M, \mathcal{E})
\]
As the examples in $\mathcal{E}$ did not contain this specific type of historical hate imagery, the model showed overfitting to their superficial patterns, defaulted to a naive visual description, and failed to recognize the unambiguous hate. This demonstrates that for certain models, non-representative few-shot examples can be more detrimental than a zero-shot prompt, as they can suppress the model's ability to apply its own knowledge.


\subsubsection{Correct Classification But Flawed Reasoning}
\begin{figure}[h!]
    \centering
    \includegraphics[width=0.95\textwidth]{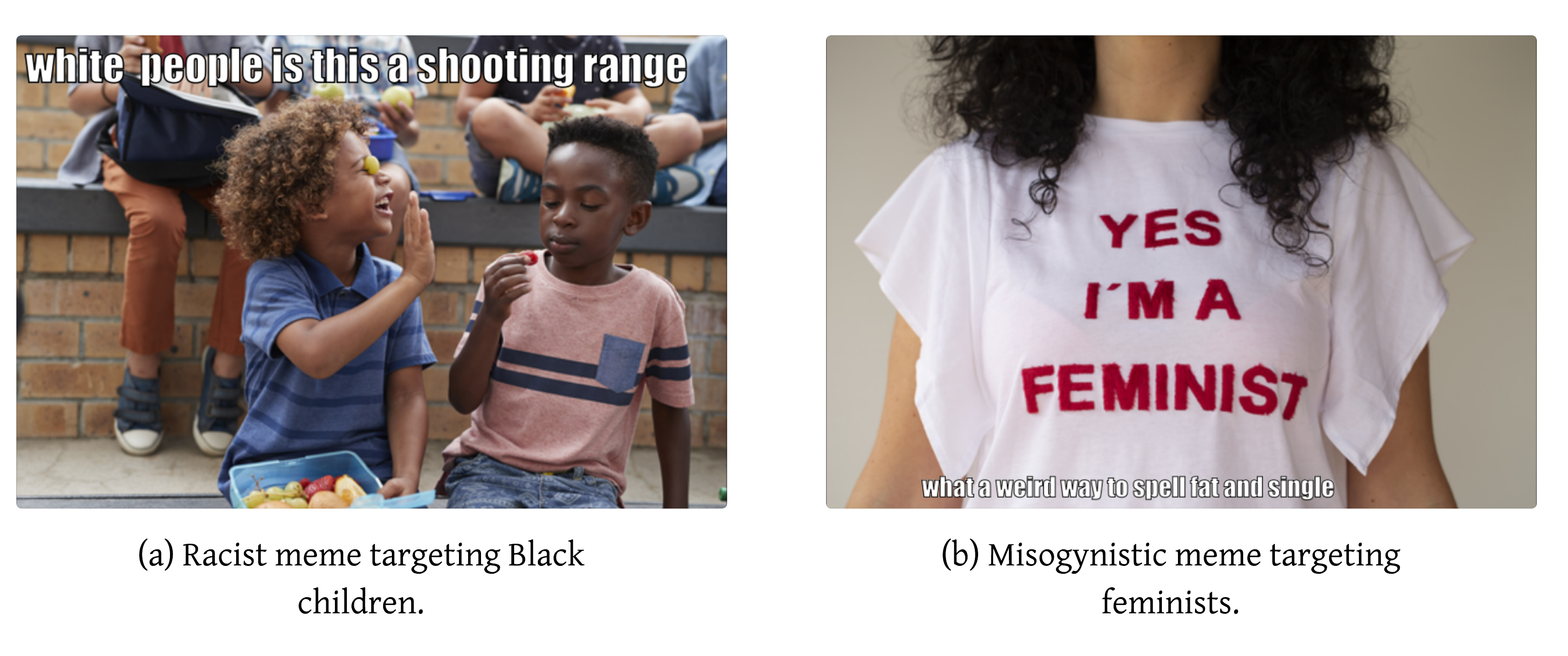}
    \caption{Hateful memes correctly classified based on flawed or hallucinated reasoning. Models here gave the correct result, but analyzing their justifications reveals their decision is based on the wrong reasons, most commonly for their tendency to focus on a benign/non-existent text template while ignoring the hateful theme.}
    \label{fig:twomemes}
\end{figure}

Instances where a model arrives at the correct classification but for the incorrect reasons are more revealing than plain mistakes. This phenomenon, often described as `getting the right answer for the wrong reason'. Figure~\ref{fig:twomemes} shows such an example. It exposes the superficial nature of the model's reasoning process and its reliance on flawed heuristics or hallucinated evidence. Such cases are particularly deceptive because a simple accuracy check would mark them as a success, masking the model's actual contextual ignorance.


\textit{Analysis:} These two cases reveal a common underlying failure: the models are not performing genuine multimodal reasoning but are instead using flawed shortcuts. We can formalize this by considering that a model's decision function $f_\theta$ should operate on the true features of a meme $M$, which is composed of its image $I$ and its actual text $T_{actual}$. The correct reasoning pathway is:
\[
\hat{y} = f_\theta(I, T_{actual}) \rightarrow \text{hateful}
\]
However, the models' justifications show they followed incorrect pathways, operating on a fabricated or incomplete feature set $M'$ where $M' \neq M$.

In the first case (Figure~\ref{fig:twomemes}a), GPT-4o-mini correctly classified the racist meme as \texttt{hateful} but justified its decision by misinterpreting the neutral meme format `white people is this X' as derogatory, completely ignoring the actual hateful phrase `is this a shooting range.' In the second case (Figure~\ref{fig:twomemes}b), LLaVA also correctly identified the misogynistic meme as \texttt{hateful} but based its justification on a hallucinated phrase (`the most dangerous people on the planet') rather than the actual text present in the image. In both instances, the model focused on a benign or non-existent text template while ignoring the hateful payload.

These cases highlight a critical challenge: models can be correct for the wrong reasons. Analyzing their justifications is therefore essential to verify true understanding and rule out `success' from flawed heuristics or hallucination.

\subsubsection{Contextual Misinterpretation}
\begin{figure}[h!]
    \centering
    \includegraphics[width=1\textwidth]{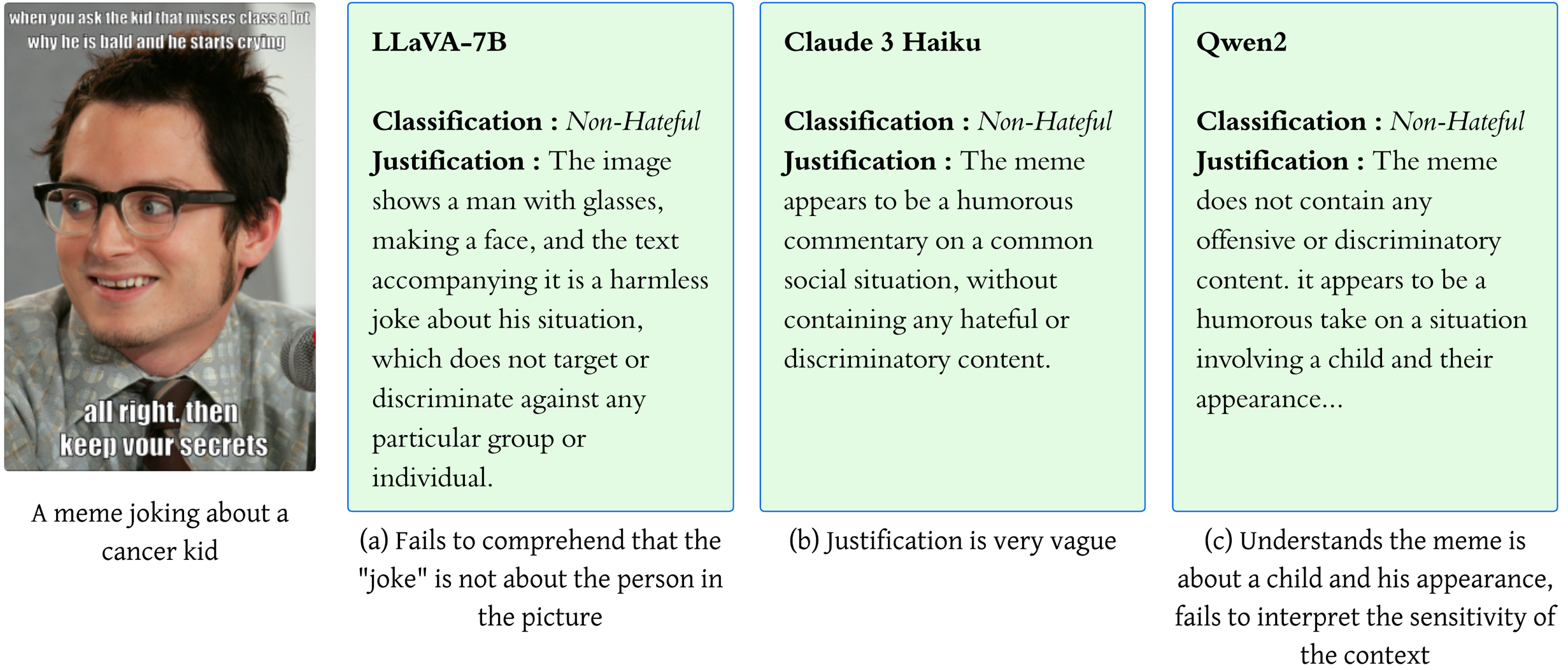} 
    \caption{Hateful Meme Requiring Real-World Inference. Models here fail to understand the context of the meme and identify the scenario as a ‘joke’, failing to access the necessary real-world knowledge.}
    \label{fig:modeljust}
\end{figure}

The incapacity of all models to infer the practical knowledge required to decode hateful subtext was a crucial flaw. They took the meme in Figure \ref{fig:modeljust} literally, but they were unable to make the necessary logical leap to recognize its cruel mockery of a child who is afflicted with cancer.


As shown in Figure~\ref{fig:modeljust}, every model classified the meme as non-hateful, providing justifications that indicate a surface-level reading.


\textit{Analysis}: The models' justifications reveal a shared cognitive gap: they identify the scenario as a `joke' but fail to access the necessary real-world knowledge. We can formalize this failure by defining the meme's features as having both explicit ($F_{\text{explicit}}$) and implicit ($F_{\text{implicit}}$) components. 

The models' decision-making appears limited to only the explicit features, approximating:
\[
f_\theta(M) \approx f_\theta(F_{\text{explicit}})
\]
whereas a successful function must operate on both. This inability to integrate commonsense knowledge is a critical limitation.

\subsubsection{Over-sensitivity to Keywords}



\begin{figure}[h!]
    \centering
    \includegraphics[width=0.95\textwidth]{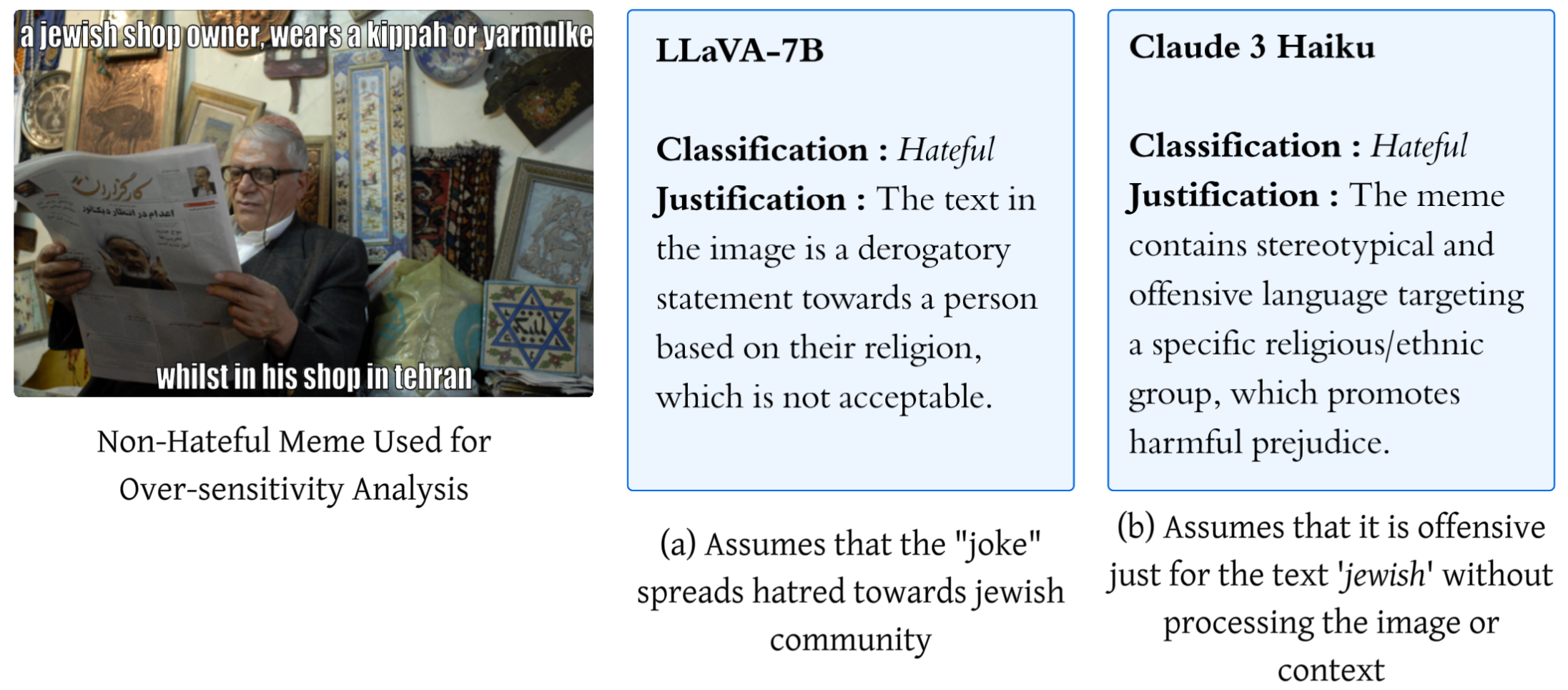} 
    \caption{Justifications for Benign Meme, models misinterpreted harmless content as hateful due to the presence of some sensitive terms involving race, religion, and others.}
    \label{fig:meme02364}
\end{figure}

A key source of false positives was \textbf{Keyword Over-Sensitivity}, where models flagged benign content as hateful due to the presence of a sensitive term. For instance, the neutral meme in Figure~\ref{fig:meme02364} was misclassified by both LLaVA-7B and Claude 3 Haiku on this basis.

The justifications provided by the models, shown in Figure~\ref{fig:meme02364}, reveal that their decisions were triggered almost exclusively by the presence of the word `Jewish'.



\textit{Analysis}: This case demonstrates a critical failure where both models hallucinated derogatory language. The models' decision function, $f_\theta(M)$, appears to have been disproportionately weighted by a single sensitive keyword, $w_s$, such that:
\[
f_\theta(M) \approx f_\theta(w_s)
\]
This suggests the models' strong safety alignment for the \texttt{keyword} overrode the neutral context of the meme, causing a false positive.


\subsubsection{Failure to Detect Coded or Nuanced Hate}

\begin{figure}[h!]
    \centering
    \includegraphics[width=1\textwidth]{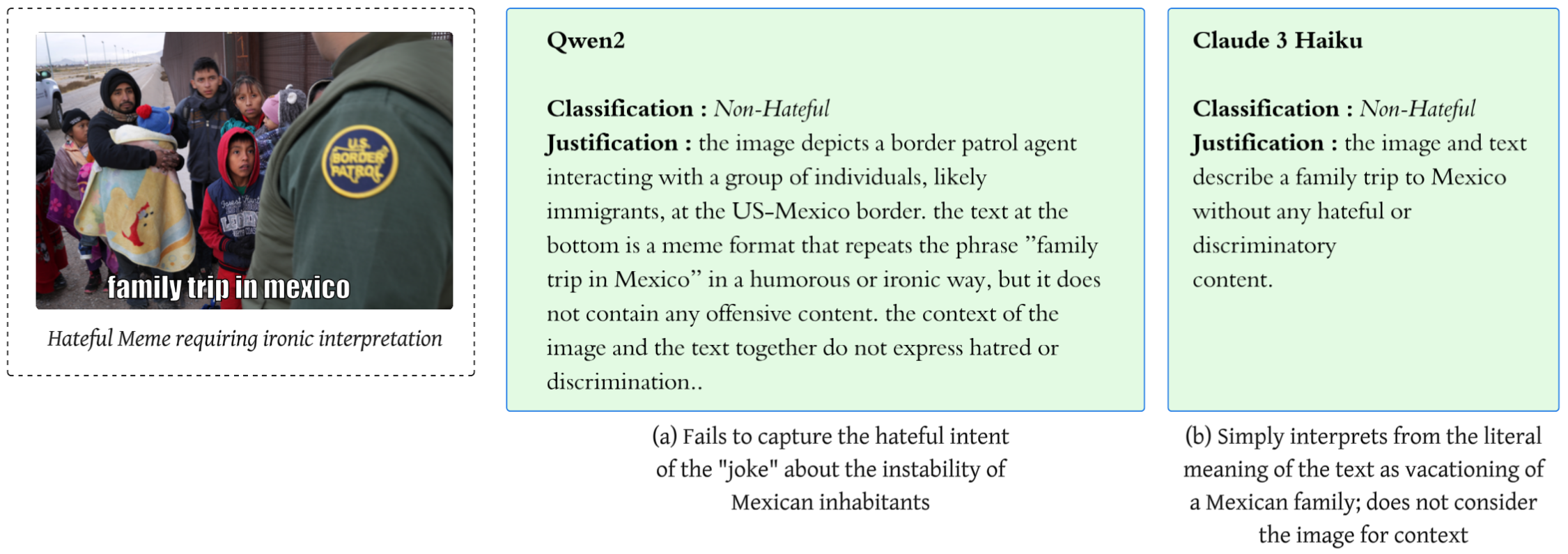} 
    \caption{Failure Case: Inter-Model Comparison on Coded or Nuanced Hate. Both Qwen2-7B and Claude 3 Haiku failed to detect the subtle hateful intent in this meme, highlighting limitations in interpreting implicit or ironic content.}
    \label{fig:meme68459}
\end{figure}

A key limitation observed was the models' failure to detect nuanced hate that requires external, real-world knowledge, a challenge central to the HMCD's design. As shown in Figure~\ref{fig:meme68459}, models performed a literal analysis but missed the hateful subtext. The cruel irony of pairing the text `family trip in Mexico' with a distressing image of an immigrant family was lost on both Qwen2 and Claude 3 Haiku, who incorrectly classified the meme as non-hateful.

\textit{Analysis}: A notable cognitive gap is revealed by this case study, where models carried out a literal analysis, recognizing the term `family trip' but not deducing from the visual cues the emotional context of distress. This failure can be formalized by defining the meme's features as having explicit ($F_{\text{explicit}}$) and implicit ($\mathcal{C}_{\text{context}}$) components. Models' decision-making was limited to the explicit features, approximating:
\[
f_\theta(M) \approx f_\theta(F_{\text{explicit}})
\]
whereas a successful function must recognize the ironic dissonance between them. This inability to move beyond literal interpretation to understand that the meme's `humor' is a form of dehumanization highlights a key frontier for VLM development.

\subsection{Quantitative Performance Overview}

To form an empirical baseline, we evaluated the four VLMs on our balanced test set using both zero-shot and few-shot prompting strategies. We used standard classification metrics, such as Precision, Recall, and F1-Score, to evaluate how well each model worked, for both \texttt{hateful} and \texttt{non-hateful} classes shown in Table~\ref {tab:quantitative_results}.

\begin{table}[h!]
    \centering
    \caption{Performance comparison of VLMs on Hateful Meme Detection under zero-shot and few-shot conditions. For the critical `hateful' class, the highest and lowest scores in each column are marked in bold and underlined, respectively.}
    \vspace{0.5em}
    \label{tab:quantitative_results}
    \renewcommand{\arraystretch}{1.3} 
    \resizebox{\textwidth}{!}{
    \begin{tabular}{ >{\centering\arraybackslash}m{2.8cm} | c | ccc | ccc }
        \hline
        \multirow{2}{*}{\textbf{Model}} & \multirow{2}{*}{\textbf{Prompt Type}} & \multicolumn{3}{c|}{\textbf{Hateful Class}} & \multicolumn{3}{c}{\textbf{Non-Hateful Class}} \\ \cline{3-8}
        & & Precision & Recall & F1-Score & Precision & Recall & F1-Score \\ \hline
        \multirow{2}{*}{\textbf{LLaVA-7B}} & Zero-Shot & \underline{0.53} & 0.38 & 0.45 & \underline{0.52} & 0.66 & 0.58 \\ \cline{2-8}
        & Few-Shot  & 0.59 & 0.39 & 0.47 & 0.57 & 0.59 & 0.58 \\ \hline
        \multirow{2}{*}{\textbf{Qwen2-7B}} & Zero-Shot & 0.70 & 0.34 & 0.45 & 0.56 & 0.86 & \textbf{0.68} \\ \cline{2-8}
        & Few-Shot  & \textbf{0.71} & \underline{0.32} & \underline{0.44} & 0.56 & \textbf{0.87} & \textbf{0.68} \\ \hline
        \multirow{2}{*}{\textbf{Claude 3 Haiku}} & Zero-Shot & 0.59 & 0.37 & 0.46 & 0.62 & \underline{0.51} & 0.56 \\ \cline{2-8}
        & Few-Shot  & 0.58 & 0.46 & 0.51 & 0.56 & 0.67 & 0.61 \\ \hline
        \multirow{2}{*}{\textbf{GPT-4o-mini}} & Zero-Shot & 0.69 & 0.51 & 0.59 & 0.61 & 0.77 & \textbf{0.68} \\ \cline{2-8}
        & Few-Shot  & 0.68 & \textbf{0.57} & \textbf{0.62} & \textbf{0.63} & 0.74 & \textbf{0.68} \\ \hline
    \end{tabular}
    }
\end{table}

\textit{Analysis of Quantitative Results}: The quantitative data reveals significant heterogeneity in the performance profiles of the evaluated models. This may be because of being trained on different sets of data, having different architectural designs, and using different alignment procedures.



\begin{enumerate}
    \item \textbf{Overall Performance and Reasoning Quality:} The community version of GPT-4o-mini consistently was the top-performing model, achieving the highest F1-Score for the `hateful' class in both prompt configurations (0.59 and 0.62). Its superior performance showed a proper balance between precision and recall. This suggests that its architecture is better suited for complex and inferential reasoning. This finding is consistent with recent studies that show the official GPT-4o model has demonstrated state-of-the-art accuracy at detecting multimodal hate speech \cite{van2025detecting}. We posit that its pre-training on a large and varied set of internet data, which contains a lot of different cultural settings and meme formats, is a good fit for the specific problems that the HMCD dataset presents.

    \item \textbf{The Precision-Recall Trade-off:} A clear trade-off emerges among the open-source models. \textbf{Qwen2-7B} achieved the highest precision (0.70 and 0.71), likely due to strong OCR and factual grounding, making it reliable for text-based hate. However, its recall was the lowest (0.34 and 0.32), missing roughly 70\% of hateful content. This statistical profile is directly supported by our qualitative findings. For example, in our analysis of nuanced hate (Figure~\ref{fig:meme68459}), Qwen2-7B performed a perfect literal analysis of the `family trip in Mexico' meme, but completely failed to grasp the cruel, ironic subtext required to identify it as hateful. This tendency to miss non-explicit hate in ambiguous cases is the primary driver of its low recall score. 

\end{enumerate}

\subsection{Discussion}
Our results highlight a clear gap between how well the models seem to understand the task and how they actually perform. Even the well-performed model, GPT-4o mini, did not always rely on solid reasoning to achieve its high scores. Instead, all models frequently repeated the same types of mistakes—often using flawed logic to land on the right answer or failing to incorporate real-world knowledge. This pattern points to a dependence on shallow strategies rather than genuine understanding. For example, Qwen2-7B showed high precision but very low recall, meaning it avoided errors by taking an overly literal approach, but in doing so, missed most subtle cases of hate. These findings suggest that accuracy alone is not a reliable measure of a model’s true ability to moderate content.


\section{Conclusion \& Future Scopes}
Our quantitative findings show that, despite recent VLMs demonstrating impressive progress, performance discrepancies persist across architectures. GPT-4o-mini exhibited the most balanced performance. More importantly, our qualitative analysis reveals that all tested models have significant vulnerabilities, limiting reliability for real-world content moderation. A key limitation is their consistent failure to infer unstated, real-world knowledge, with reasoning largely relying on the literal meaning of images and text and missing essential common-sense and emotional understanding. While VLMs are promising tools, they are not yet capable of reliably identifying complex multimodal hate speech, lacking the deep inferential skills necessary to interpret subtle expressions of hate and remaining fragile under contextual variations.

Future research can focus on targeted evaluations and enhancing reasoning capabilities. Extending studies to alternative prompting strategies, larger model families, and multilingual datasets can reveal whether observed failure modes are language-specific or universal. Assessing performance on specialized datasets, such as MAMI, can help determine whether models can identify precise targets and sub-categories of hate. Additionally, developing mitigation strategies can directly address observed error archetypes, such as `Example-Induced Bias' and `Flawed Reasoning'. Collectively, these directions offer a path toward more reliable VLMs for content moderation and robust multimodal reasoning.

\let\cleardoublepage\clearpage